# Unveiling Intractable Epileptogenic Brain Networks with Deep Learning Algorithms: A Novel and Comprehensive Framework for Scalable Seizure Prediction with Unimodal Neuroimaging Data in Pediatric Patients

Bliss Singhal[1], Fnu Pooja[2]
[1]Bellevue College, Washington, [2]Google, Washington, +poojasn@google.com

**Abstract**
*Epilepsy is a prevalent neurological disorder affecting approximately 50 million individuals worldwide and 1.2 million Americans. There exist millions of pediatric patients with intractable epilepsy, a condition in which seizures fail to come under control. The occurrence of seizures can result in physical injury, disorientation, unconsciousness, and additional symptoms that could impede children's ability to participate in everyday tasks. Predicting seizures can help parents and healthcare providers take precautions, prevent risky situations, and mentally prepare children to minimize anxiety and nervousness associated with the uncertainty of a seizure. This research proposes a novel and comprehensive framework to predict seizures in pediatric patients by evaluating machine learning algorithms on unimodal neuroimaging data consisting of electroencephalogram signals. The bandpass filtering and independent component analysis proved to be effective in reducing the noise and artifacts from the dataset. Various machine learning algorithms' performance is evaluated on important metrics such as accuracy, precision, specificity, sensitivity, F1 score and MCC. The results show that the deep learning algorithms are more successful in predicting seizures than logistic Regression, and k nearest neighbors. The recurrent neural network (RNN) gave the highest precision and F1 Score, long short-term memory (LSTM) outperformed RNN in accuracy and convolutional neural network (CNN) resulted in the highest Specificity. This research has significant implications for healthcare providers in proactively managing seizure occurrence in pediatric patients, potentially transforming clinical practices, and improving pediatric care.*

**Keywords:** *Intractable epilepsy, seizure, deep learning, prediction, electroencephalogram channels.*

## 1. Introduction

Intractable Epilepsy is a condition where a patient's seizures cannot be controlled with treatment [1]. Seizures can result in physical harm, loss of consciousness, confusion, and can make activities such as driving or swimming dangerous. Out of 50 million cases of epilepsy worldwide [2], 25% of diagnoses are in children, and 30% of these cases are intractable epilepsy [3], [4]. Current treatment options for intractable epilepsy, such as surgery or neurostimulation therapy, are invasive and pose potential risks [5]. Individuals, particularly children, who suffer from intractable epilepsy are at a higher risk of experiencing emotional and behavioral issues, which can have a negative impact on their quality of life [6]. Predicting seizure attacks can have several benefits. Firstly, seizures can significantly impact a child's quality of life and limit their ability to perform daily activities such as playing, attending school, or swimming. Predicting seizure attacks can improve the quality of life for people, especially children with intractable epilepsy, help clinicians optimize treatment plans, and advance our understanding of the condition.

Overcoming societal stigma and discrimination surrounding epilepsy is a significant challenge faced by children with epilepsy and their families. Since intractable epilepsy cannot be cured or fully controlled, predicting an impending seizure can help parents and caregivers monitor children and reduce the risk of seizures occurring in unsupervised environments. Awareness of the possibility of a seizure attack can also better prepare a child mentally and decrease anxiety and nervousness compared to not knowing when the next seizure might occur.

### 1.1 Seizures: An Overview
Epilepsy is a rapid and early abnormality in the brain's electrical activity, disrupting part or all the human body. Medical researchers have divided epileptic seizures into three categories: generalized, focal, and epilepsy with unknown onset. Focal epilepsy involves seizures that begin on one side of the brain or involve one area. General epilepsy is a seizure that affects all areas of the brain. The onset area of the seizure may be vague or difficult to exactly recognize its location, in which case the seizure belongs to an unknown group.

### 1.2 Available Medical treatments and their limitations
Epilepsy is treated using various methods, such as medications, surgery, devices, and sometimes diet. Anti-seizure drugs are the main treatment method for epilepsy, with doctors prescribing drugs like Brivaracetam (Briviact), Cannabidiol (Epidiolex), Carbamazepine (Tegretol), Cenobamate (Xcopri), Clonazepam (Klonopin), and Clobazam (Onfi) based on the type of seizure. If the initial medication does not work, doctors may switch to another or add more medication. In specific emergency situations, there are rescue medications and treatments that can help stop a seizure quickly. These medications can be given nasally, orally, sublingually, buccally, or rectally, depending on the circumstance. The most commonly used medications are benzodiazepines, including Diazepam, Valtoco (nasal spray), Diastat (rectal), Lorazepam, and Midazolam (Nayzilam, buccally or orally). Anti-seizure medications can effectively control seizures in many people with epilepsy. However, they may not work for everyone, and some people may experience side effects such as dizziness, fatigue, or nausea. Additionally, if a person is taking multiple medications, there may be a risk of drug interactions that can decrease the effectiveness of the medications or cause additional side effects [7].

Surgery could be an option if medication fails to control seizures or if seizures are caused by brain problems such as a tumor or stroke. During surgery, the doctor removes a small part of the brain causing seizures or makes small cuts to prevent seizures from spreading. However, surgery carries the risk of complications such as infection or bleeding, and there is also a risk of cognitive or motor deficits after the surgery. Devices such as Vagus nerve stimulation (VNS) or Responsive neurostimulation (RNS) are also approved to treat epilepsy [8]. Finally, a ketogenic diet, which is high in fat and low in carbs, is used to control seizures in children and might work for adults, although more research is needed [9]. The diet is strict and complicated, so patients must work closely with their doctors.

For people with intractable epilepsy, treatment options are limited, and existing options such as surgery or neurostimulation therapy are invasive, complicated, and pose potential risks. Individuals with intractable epilepsy may also be at a higher risk of experiencing emotional and behavioral issues, and the resulting psychological impact can have a negative effect on their quality of life. Overall, while the available methods to treat epilepsy have shown promise, there is still much that is not fully understood about the condition, and more research is needed to develop more effective treatments with fewer side effects and risks.

### 1.3 Importance of Predicting Seizures
Predicting seizure attacks can be important for several reasons. Firstly, seizures can significantly impact a child's quality of life and limit their ability to perform daily activities. Seizures can cause physical harm, loss of consciousness, confusion, and other symptoms that may make it difficult or dangerous for children to engage in daily activities such as playing, attending school, or swimming. Predicting when seizures may occur, can allow parents, and healthcare providers to take precautions and avoid potentially hazardous situations for children suffering from intractable epilepsy. Knowing when a seizure is likely to occur will assist as well to reduce stress and anxiety for children with epilepsy as well as for parents, and they will be able to prepare for the occurrence of the seizure. Additionally, knowing when a seizure is likely to occur will open up more opportunities and activities that children suffering from intractable epilepsy can partake in, such as swimming, sports, and other school and recreational activities. To discuss more about the practicality of predicting seizures from electroencephalogram data, electroencephalograms are in wide use in hospitals and clinics, and this is only going to rise in the coming years. In 2014, 28% of hospitals included in the National Inpatient Sample, with 11.6% of these hospitals using continuous electroencephalograms. The proportion of hospitals using electroencephalograms increased by 122.7% between 2003 and 2014 [10]. As more research is done involving electroencephalograms, their benefits for diagnosing and treating patients will only increase, and subsequently, their availability in hospitals will also increase. Furthermore, in 2018, Hospitals dominated the market for electroencephalogram devices with it being approximately 70% of the market for electroencephalogram devices. As a result, hospitals and clinics equipped with electroencephalogram machines would be accessible to a large proportion of pediatric patients with intractable epilepsy, enabling them to obtain electroencephalogram recordings. Subsequently, healthcare providers can utilize a seizure prediction model to forecast potential seizures based on the data obtained from these recordings, allowing for preventative measures to be taken to manage the condition. Additionally, predicting epilepsy attacks can also help researchers better understand the condition and develop more effective treatments for children. By studying patterns in brain activity and other factors leading up to seizures, researchers may be able to identify new targets for therapy and improve our understanding of the mechanisms underlying epilepsy. Overall, predicting epilepsy attacks can improve the quality of life for people especially children with intractable epilepsy and advance our understanding of the condition.

### 1.4 Social Impact
Overcoming the societal stigma and discrimination surrounding epilepsy is a significant challenge that is faced by children with epilepsy and their families. Since intractable epilepsy cannot be cured or fully controlled, the ability to predict an impending seizure can help parents and caregivers monitor children and reduce the risk of seizures occurring in unsupervised environments. Awareness of the possibility of a seizure attack can better prepare a child mentally and can decrease the anxiety and nervousness compared to not knowing when the next seizure might occur. Thus, this research has the potential to significantly enhance the quality of life for children with intractable epilepsy and help in reducing the stigma attached with this disease. Moreover, using classification algorithms to detect seizures can allow children with intractable epilepsy to engage in essential activities such as attending school, playing outdoors, or swimming, which they may have previously avoided due to the risk of experiencing seizures at any moment. By analyzing large amounts of data from brain electrical waves and other sources, these classification algorithms can significantly improve the prediction of seizures by identifying patterns that are difficult or impossible for humans to detect. This identification can lead to earlier, as well as more effective treatments. Seizure medications are frequently prescribed at high doses to prevent seizures, but this approach can result in undesirable side effects, particularly in children. However, by predicting the likelihood of a seizure occurrence, medications can be administered only when necessary, thereby reducing the overall occurrence of side effects in children.

### 1.5 Past Research
Past research includes many various methods of predicting seizures, with the use of many different preprocessing and classification algorithms, including principal component analysis (PCA), wavelet transforms, support vector machines (SVM), and random forests. However, past research's prediction time is usually within a few seconds or minutes of the seizure, which is not a practical amount of time given to prepare for the seizure attack. Additionally, majority of these research's solutions are tailored towards adults. This poses a problem as its applicability to pediatric patients would be limited because of this and therefore, would not be suitable for these segments of patients. There are a few proposed methods specifically for pediatric patients, but their prediction times are either only a few seconds or minutes or have low performance. For example, Yang *et. al.* (2015) proposed a seizure prediction method for pediatric patients with a sensitivity and specificity of only 59% and 81%, respectively. Additionally, Behnam *et al.* (2016) and Zhang *et. al.* (2016) proposed solutions with a prediction time of only 6.64 seconds and 2 seconds, respectively, which would be not enough time before for the patient to adequately prepare for the seizure [11].

### 1.6 Research Hypothesis and Engineering Goals

This study proposes a novel seizure prediction method specifically for pediatric patients with a significantly high performance as well while having a large prediction time to give pediatric patients with intractable epilepsy adequate time to prepare using machine learning. In doing so, this will allow for timely interventions and enhance these patients' quality of life. This research specifically aimed to maximize the signal to noise ratio in the dataset. This is because electroencephalogram datasets, which is what this study uses, tend to be very noisy, because electroencephalograms record brain activity, which can be influenced by many different factors, including environmental noise, physiological noise, and subject specific factors. Environmental noise refers to electrical signals in the environment from other electric devices which can interfere with the electroencephalogram recordings. Physiological noise refers to variable electrical activity in the brain that is unrelated to the focus, such as eye movements, heartbeats, and respiratory activity. Subject specific factors refer to variability that will occur due to variation in subjects. Different subjects may have different skull and brain structures, as well as different resistance in their scalp, which can affect electroencephalogram signals. As a result, it is critical to reduce noise as much as possible to obtain the highest seizure prediction performance.

Secondly, this research intended to implement a cost effective and practical system for automated seizure prediction that utilizes a minimal number of Electroencephalogram channels. Typically, in electroencephalogram devices, there is a direct correlation between the number of channels the device utilizes, and the cost of the device. The average price of a 16-channel electroencephalogram device is approximately $900, while the average price of an 8-channel electroencephalogram device is circa $500, and the average price of a 4-channel electroencephalogram device is only $200 to $300. Because of this, it is imperative that the study's seizure prediction model utilize minimal electroencephalogram channels, as this would reduce the financial barriers for clinics and hospitals to utilize this method.

This research also seeks to develop a comprehensive framework for automated seizure prediction that leverages a range of classification models designed to accurately identify complex patterns and subtle changes in Electroencephalogram data that indicates the future occurrence of a seizure. We wanted to compare the performance of deep learning algorithms, which tend to be quite complex, to that of more simple and standard classification algorithms. In doing so, we wanted to see that in the context of electroencephalogram data, which tends to have subtle and complex patterns as well as a lot of noise and artifacts, would more complex or more simple algorithms be more suited for analyzing this data. Complex algorithms may overfit data due to the data's abundance of distortion, while simple algorithms may overlook the subtle trends in the electroencephalogram data that forecast seizure occurrence.

The last objective of this study is to conduct a robust comparative study of the prediction accuracies achieved by different machine learning models when predicting seizures at various time horizons before the seizure onset. In doing so, we are able to analyze the maximum time period before which the seizure prediction models can be used for as well as any trends or differences that occur in the quality of the prediction model over time.

### 2. Methodology

This research was conducted in multiple phases, with each subsequent phase dependent on the successful execution of the previous one. The results obtained from each phase were utilized as input for the subsequent phase.

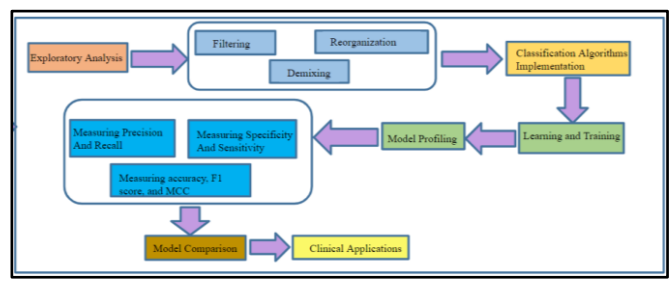

*Figure 1: Comprehensive Research Framework*

### 2.1. Dataset and Data Analysis

This research utilizes CHB-MIT Scalp brain electrical recordings obtained from the Children Hospital Boston (CHB) and the Massachusetts Institute of Technology (MIT) [12]. These recordings are from real pediatric patients suffering from intractable seizures and contain electroencephalogram signals. As the electroencephalogram datasets are complex and large in nature, significant effort was put into data analysis. To make the resultant dataset suitable for machine learning algorithms, multiple Python libraries and preprocessing algorithms were used for data segmentation, cleansing, filtering, and noise reduction.

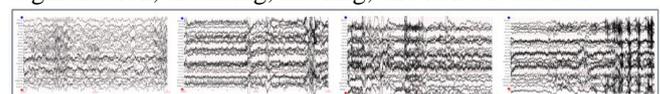

*Figure2: Electroencephalogram dataset*

Each electroencephalogram dataset contains 9-42 continuous files of electroencephalogram recordings, and each file spans approximately an hour in length. The dataset in every electroencephalogram recording has approximately 23 channels, with 1280 voltage values per channel. The signals from the electroencephalogram are captured at a resolution of 16 bits with a sampling rate of 256 samples per second and frequency from 0 Hz to 128 Hz. Accompanying these files for each of the datasets was a summary file detailing the start and end times for the files as well as for the seizures within those files as well as the Electroencephalogram channels. Many files in the datasets were duplicates, as multiple files shared the same timestamp, so these files were removed. The files in the dataset were also out of order, so they were reordered using the file start and end times provided in the summary file. The MNE database was then used to extract the data into epochs with each epoch representing five seconds worth of data. MNE-Python is an open-source software library that covers multiple methods for data preprocessing, source localization, statistical analysis, and estimation of functional connectivity between distributed brain regions. As the dataset did not have a montage configuration, the standard Electroencephalogram montage was applied to each of the datasets. To do this, multiple channels had to be deleted since they did not fit into the montage, resulting in 16 channels. Additionally, the channels were not in the correct format for the montage and had to be renamed.

The labels for this dataset detailing when a seizure is occurring also needed to be made. This research would be a binary classification problem, so the labels would be either one or zero, with "one" indicating that the seizure is occurring and a "zero" indicating that a seizure is not occurring. To determine when a seizure was occurring, the seizure start and stop times provided from the summary file was used. However, the seizure times were localized. For example, even though the file time was t = 3600 seconds to t = 7200 seconds, the seizure time within that file would be t = 20 seconds to t = 80 seconds. These timings were then standardized to be in the context of the entire dataset, so the seizure start and stop times would therefore then be t = 3620 seconds to t = 7280 seconds. If an epoch fell into the time intervals during which a seizure occurred, they would receive a label of one, otherwise, they received a label of zero. Any of the epochs and labels that occurred after the last seizure were deleted since they were unnecessary information. It is important to note that labels had to be compiled for each out of the 23 datasets individually, since for each dataset, the timestamp was reset to zero since it was for a different patient.

All the epochs were then combined into one epochs object, and all the labels were compiled into one array as well. The resulting epoch dataset of 230,000 epoch samples is divided into a training set containing 200,000 epoch samples, and a testing dataset containing 30,000 epoch samples for the classification algorithms.

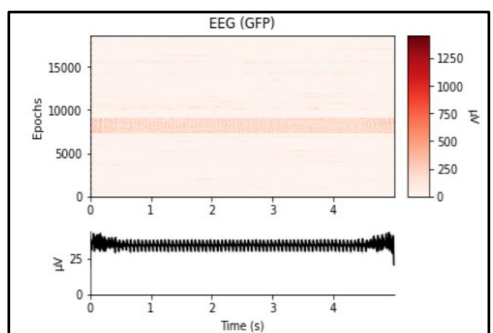

*Figure 3: Epoch converted from an electroencephalogram signal*

## 2.2 Machine Learning

These datasets are used as an input in five machine learning pipelines where the first three pipelines consist of deep learning algorithms: Long short-term memory (LSTM), recurrent neural network (RNN), and convolutional neural network (CNN), and the last two pipelines are made up of logistic regression (LR), k-nearest neighbors (k-NN).

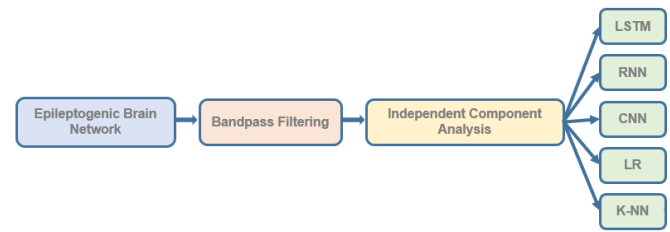

*Figure 4: Machine Learning algorithm pipeline used in the research*

### 2.2.1 Preprocessing Methods

Preprocessing methods can help to reduce the noise and artifacts in the data, improve signal quality, and increase the sensitivity of the analyses. This can lead to more accurate and meaningful results and improve our understanding of the underlying neural processes in various brain disorders, such as epilepsy. Two preprocessing algorithms are used for filtering, demixing, and noise reduction.

The very first preprocessing step is performed by using the bandpass filtering algorithm. This algorithm performs both low-pass and high-pass filtering by removing values below and above a certain frequency threshold, respectively. It is applied to all epochs individually where each epoch has various characteristics such as number of epochs, number of channels, and number of time values. Frequencies below 1 Hz are typically pulse artifacts and other low-frequency noise that can come from vibrations in the building or nearby electromagnetic fields. Frequencies above 40 Hz are also typically not important as they consist of involuntary eye movement and are vulnerable to interference by lamps or other devices. Prior research [13], [14] has shown that seizures typically occur between frequencies of 3 Hz to 30 Hz and from 40 Hz to 50 Hz. Frequencies below 1 Hz, and above 50 Hz are filtered out from the epoch dataset.

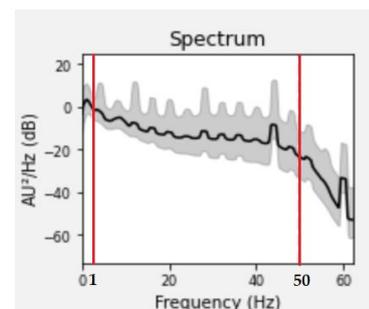

*Figure 5: Graph of epochs in one dataset depicting the application of band pass filter*

Independent component analysis (ICA) was the second preprocessing step which was aimed at further separating the artifacts from neuronal components and reducing the number of components where both are mixed as well as reducing the number of channels in the dataset. Independent component analysis is a demixing algorithm that isolates statistically independent signals. ICA is especially important since it can be used to isolate a particular behavior or pattern in the Electroencephalogram data. ICA is focused on channels and reframes these channels into these independent components [15]. The specific ICA algorithm that was used was FastICA. We selected 10 ICA components. The resulting dataset now had 10 features, and the shape of each epoch was now (10, 1280) instead of (16, 1280).

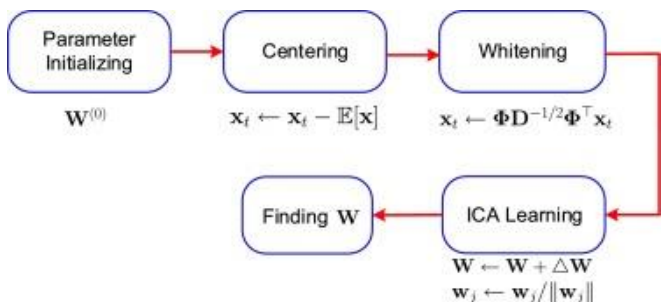

*Figure 6: Diagram showing the process of independent component analysis*

### 2.2.2 Algorithms
Five classification algorithms are used to predict seizure in brain electroencephalogram waves. These are logistic regression (LR), k-nearest neighbors (k-NN), and three deep learning algorithms: long short-term memory (LSTM) neural network, recurrent neural network (RNN), and convolutional neural network (CNN).

| Logistic Regression | A simple and efficient algorithm designed primarily for classification problems, and most specifically towards binary classification. Since logistic regression is a form of linear regression, it tries to find a linear relationship between the components, or independent variables in the training data and the output value. Logistic regression uses a sigmoid function to find the relationship between the training data components and the labels [16]. |
|---|---|
| k-Nearest Neighbors | An algorithm that classifies the testing data based on its similarity to the training data. It is considered a lazy learner because there is no training phase necessary, for it straightaway compares the testing data to the training data without needing to learn from the training data itself. The k-nearest neighbors algorithm examines each testing case and compares each of the component's data values to that of the training cases. It calculates the similarity for each one using the distance formula for finding the distance between the two values using the Euclidean distance formula [17]. |
| Convolutional neural networks (CNNs) | A powerful tool for processing data with a grid-like topology, and their ability to automatically learn features from raw data makes them well-suited for a wide range of applications. It has the ability to identify spatial and temporal patterns in the input data by applying a series of convolutional and pooling layers [18]. |
| Recurrent neural networks (RNNs) | An algorithm for processing sequential data, and has the ability to maintain memory of previous inputs makes them well-suited for my epochs dataset. RNN allows information to be passed from one step of the sequence to the next. It uses feedback connections that allow it to use its own output as input for the next step in the sequence. The key characteristic of an RNN is its ability to maintain a state or memory of previous inputs, which allows it to make predictions based on the current input and the context provided by the previous inputs [19]. |
| Long short-term memory networks (LSTMs) | A type of RNN that is specifically used for handling long term dependencies in sequential data, such as time series data. It utilizes several gates to control the flow of information within the memory cell. LSTMs are able to handle the vanishing gradient problem encountered by traditional RNNs, where the gradient decreases to a miniscule amount while training, inhibiting the algorithm being able to find the optimal performance [20]. |

*Table1: Definition of classification algorithms*

## 3. Results and Discussion
### 3.1 Prediction Model's performances
These prediction models were tested on multiple time intervals: 20 minutes, 40 minutes, 60 minutes, 80 minutes, 100 minutes, 2 hours, 4 hours, 6 hours, 8 hours, 10 hours and 12 hours. This means that the model would provide a prediction that the seizure would occur these many minutes/hours after the data provided. The performance of the prediction models was evaluated based on 6 different metrics: precision, accuracy, sensitivity, specificity, F1 score, and Maxwell's Correlation Coefficient.

$$\text{Precision} = \frac{\text{Number of True Positives}}{(\text{Number of True Positives} + \text{Number of False Positives})}$$

$$\text{Accuracy} = \frac{\text{True Negatives} + \text{True Positives}}{(\text{True Positives} + \text{False Positives} + \text{True Negatives} + \text{False Negatives})}$$

$$\text{Specificity} = \frac{\text{Number of True Negatives}}{(\text{Number of True Negatives} + \text{Number of False Positives})}$$

$$\text{Sensitivity} = \frac{\text{Number of True Positives}}{(\text{Number of True Positives} + \text{Number of False Negatives})}$$

$$\text{F1 Score} = \frac{2 * \text{Precision} * \text{Recall}}{(\text{Precision} + \text{Recall})}$$

$$\text{MCC} = \frac{(TP * TN) - (FP * FN)}{\text{Sq Root }[(TP + FP)(TP + FN)(TN + FP)(TN + FN)]}$$

*Figure 7: Mathematical formula for metrics*

These metrics illustrated different aspects of the performance of these models and provided valuable insights.

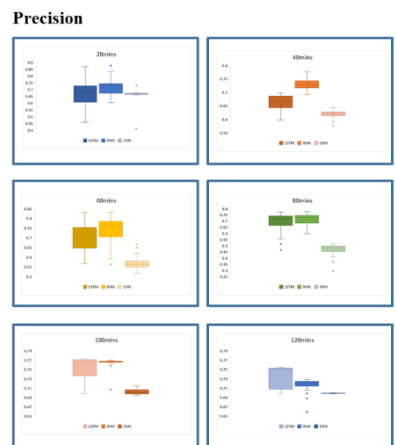

*Figure 8: Boxplots showing the precision scores for training iterations of LSTMs, RNNs, and CNNs*

These figures show that LSTMs had more variability in its precision metrics for different iterations when compared to RNNs and CNNs. However, LSTM models had the least outliers, only having two in the 40-minute time interval. This means that while LSTM models have more variability, they are more reliant and consistent in their precision metric. Also, LSTMs and RNNs seem to be alternating in which has the maximum precision value. While CNN seemed to have a
5 | P a g e

similar median to that of RNNs and LSTMs in the 20-minute time interval plot, its overall precision seems to be much lower for subsequent time intervals. CNNs also had the least variability in its precision values when compared to LSTMs and RNNs. This indicates that models that had more variability in its results also had higher precision scores.

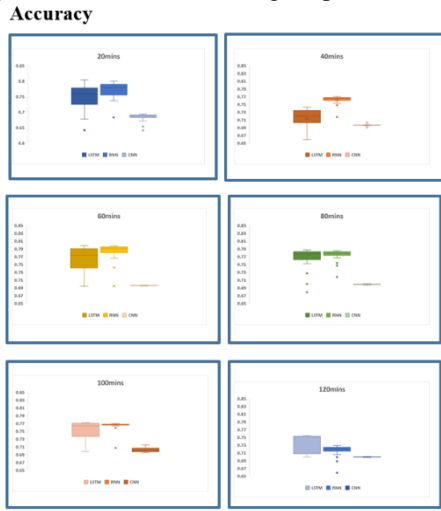

*Figure 9: Boxplots showing the accuracy scores for training iterations of LSTMs, RNNs, and CNNs*

These results are extremely similar to that of precision metrics, the only difference being that LSTMs seem to be performing the highest for most of the time intervals. There are also significantly more outliers present for these plots compared to that of precision. Additionally, all the plots seem heavily left skewed. The top 25-50% of the accuracy scores seem to be quite consistent for each of the models, while the bottom 25-50% of the accuracy scores seem to have large variation. This means that these models can stay at high accuracies for multiple iterations of the models. All the outliers also seem to be at extremely low accuracy scores, which may indicate the initial training iterations of the models. Since the variation in the accuracies of the CNN are so minimal, this indicates that the number of training iterations done has little impact on the CNN's accuracy. To compare with logistic regression and K Nearest Neighbors, these deep learning algorithms had significantly higher performance than that of logistic regression and K Nearest Neighbors, as logistic regression and K Nearest Neighbors highest accuracy overall was only 0.70 and 0.69, respectively.

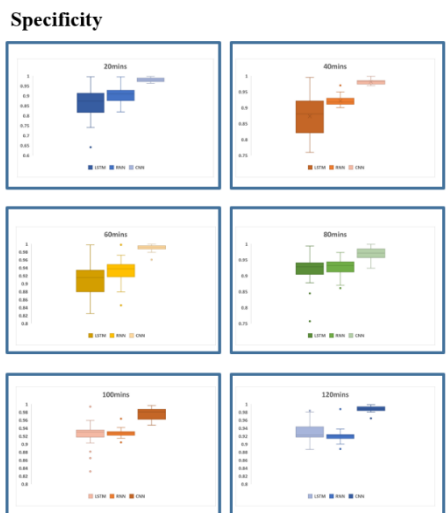

*Figure 10: Boxplots showing the specificity scores for training iterations of LSTMs, RNNs, and CNNs*

In these plots, CNN consistently has higher specificity than LSTMs and RNNs. However, some iterations of LSTMs and RNNs are able to reach to specificity levels very similar to that of CNNs, demonstrating LSTMs and RNNs potential to achieve high specificity scores as well. LSTMs have much more outliers for specificity, indicating that that the specificity scores are extremely low for multiple iterations. Looking across different time intervals, CNN's scores are consistent, while the LSTMs variation seems to be quite variable, having high variation across the 20, 40, and 60 minute time intervals, while having much lower variation for the 80, 100, and 120 minute time intervals. RNNs however, seem to have consistent variation across time intervals, as well as having lower variation than LSTMs. While the middle 50% of specificity scores are quite similar for both LSTM and RNN, LSTM is able to reach higher specificity levels than RNN, indicating that LSTM delivers greater specificity than RNN.

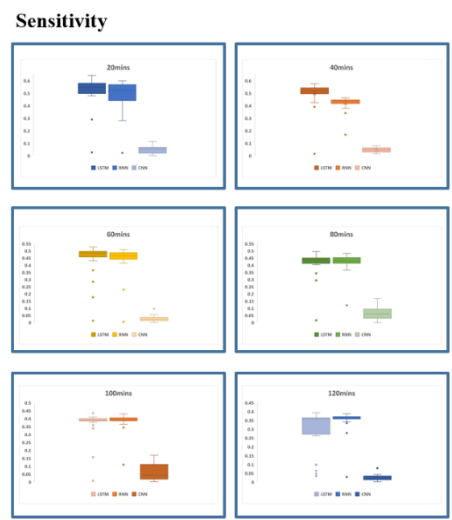

*Figure 11: Boxplots showing the sensitivity scores for training iterations of LSTMs, RNNs, and CNNs*

LSTMs and RNNs are having greater number of outliers for the distribution of sensitivity. CNN is consistently having the lowest sensitivity score, with even its highest sensitivity score being lower than that the lowest sensitivity scores of LSTMs and RNNs. Additionally, RNNs seem to have much lower sensitivity scores for the 20 and 40 minute time intervals, yet it's able to achieve similar scores to the LSTM model from the 60 minute time interval onwards. This shows that RNNs have similar performance based on sensitivity when compared to the LSTM model from time intervals 60 to 120 minutes, while LSTM models have better performance for the 20 and 40 minute time intervals. Most of the box plots have roughly symmetric distributions as well, meaning that the sensitivity scores are symmetrically distributed throughout the iterations, suggesting that the number of iterations did have a significant impact on the sensitivity scores. Compared to other metrics, LSTMs have low and decreasing variation in its sensitivity scores across all the time intervals except the 120 minutes time interval. This means that for each subsequent time interval up to 100 minutes, the LSTM model is able to have more consistent sensitivity scores.

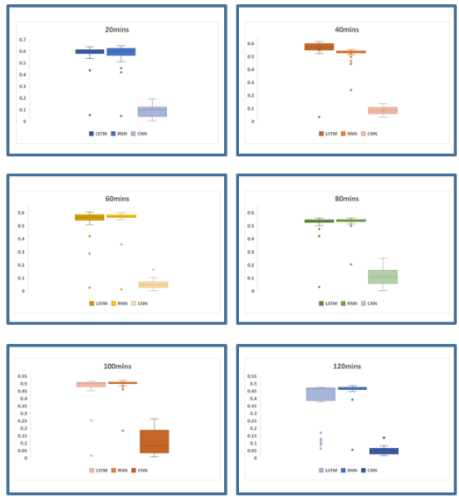

*Figure 12: Boxplots showing the F1 score for training iterations of LSTMs, RNNs, and CNNs*

RNNs and LSTMs seem to have similar performances for this metric, but CNN's score is again much lower. For some time intervals, LSTM's maximum F1 score is slightly higher than that of RNN, while for other time intervals, RNN's maximum F1 score is higher than that of LSTM. The range and interquartile range in all the models is low, indicating low variation, but they have significant outliers. This means that the majority of F1 scores are consistent, but a select few are much lower. Towards larger time intervals, both CNN and LSTM have greater variation, while RNN's variation stays consistently low, indicating that RNN gives more consistent F1 scores across all time intervals.

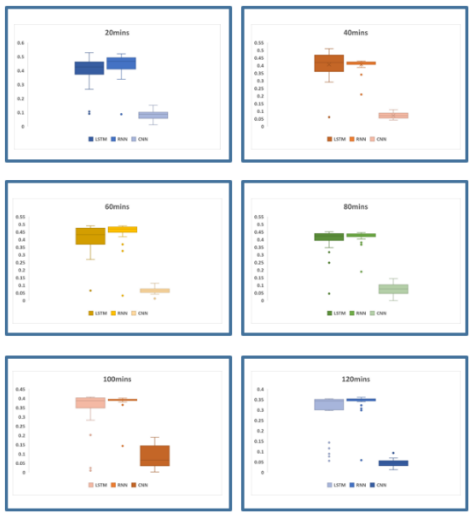

*Figure 13: Boxplots showing the MCC for training iterations of LSTMs, RNNs, and CNNs*

The MCC scores in general were lower than that of other metrics. This could be because that MCC only gives high scores if the models obtained exceptional results in all four categories: true positives, true negatives, false negatives, and false positives. This is not quite feasible giving the amount of noise and artifacts present in electroencephalogram data, which is why these models had low scores. This may also explain the skewness present in many of the boxplots which also indicate more unpredictability of MCC scores of these models. LSTM has significant variation in its MCC scores for all the time intervals, which is similar to its score distribution in accuracy and prediction. RNNs had the lowest variation in its MCC scores for all the time intervals. As the time interval increased, the variation in RNN's scores decreased, while LSTM's variation stayed fairly consistent, and CNN's variation increased. CNNs had much lower MCC scores compared to LSTM and RNNs for all the time intervals. Along with its scores in other metrics, this indicates that CNNs are not suitable for seizure prediction in this study. Except for the 40 minute time interval where LSTM had a significantly higher MCC score, LSTMs and RNNs had similar MCC scores for the other time intervals, indicating that they both have similar performance based on MCC. However, as RNNs have less variation, they could have more consistent performance.

| Prediction Models | Precision | Accuracy | Specificity | Sensitivity | F1 Score | MCC |
|---|---|---|---|---|---|---|
| 20min | 0.8763 (RNN) | 0.804 (LSTM) | 0.9991 (CNN) | 0.6419 (LSTM) | 0.6464 (RNN) | 0.5278 (LSTM) |
| 40min | 0.8618 (RNN) | 0.8012 (LSTM) | 0.9984 (CNN) | 0.5771 (LSTM) | 0.6226 (RNN) | 0.5102 (LSTM) |
| 60min | 0.8307 (LSTM) | 0.7977 (LSTM) | 0.9995 (CNN) | 0.5272 (LSTM) | 0.6041 (LSTM) | 0.4898 (RNN) |
| 80min | 0.7762 (RNN) | 0.7872 (LSTM) | 0.9988 (CNN) | 0.4954 (LSTM) | 0.5584 (LSTM) | 0.4507 (LSTM) |
| 100min | 0.7298 (LSTM) | 0.7725 (LSTM) | 0.9968 (CNN) | 0.4341 (LSTM) | 0.5210 (RNN) | 0.4073 (LSTM) |
| 120min | 0.6856 (LSTM) | 0.7562 (RNN) | 0.9987 (CNN) | 0.3939 (LSTM) | 0.4866 (LSTM) | 0.3612 (LSTM) |

*Figure 14: Resultant metrics for machine learning algorithms for different time intervals*

As shown in the figure above, LSTMs consistently gave the highest accuracy out of all the models. Its performance stayed quite consistent from 20 minutes to 60 minutes, declining overall by only 0.0063. However, afterwards, its performance declines by approximately 0.01 for every subsequent increase by 20 minutes in the time period and declines even steeper by approximately 0.02 when going from 100-120 minutes with RNN now giving a higher accuracy than LSTM. RNN and LSTM models gave the highest precision out of all models, but again there was an increasingly significant decline in precision as the time interval increased. Though it remained fairly consistent between 20 to 60 minutes again, declining overall by only 0.047, the precision decreased sharply each time interval after. A similar trend can be found for F1 score and MCC as well, declining by only 0.042 and 0.038 respectively from 20 minutes to 60 minutes, and declining much more sharply for the time intervals after. LSTMs and RNNs also gave the highest performances for both metrics at different time intervals. However, sensitivity and specificity did not exhibit these trends. Sensitivity declined with each subsequent time interval by a linear rate, by approximately 0.05, though similar to prior results, LSTMs gave the highest performance for sensitivity out of all the models for all the time periods. Specificity remained extremely high at over 0.996 and remained consistent across all the time intervals, minimal decline in value. This may be because of the high proportion of negatives to positives in the data that causes class imbalance which would cause there to be a high proportion of true negatives to all negative cases. In addition, CNNs gave the highest specificity value out of all the models for all the time periods. Overall, these metrics show that LSTMs and RNNs outperformed all the other models and that the optimal time interval for seizure prediction is between 20 and 60 minutes. More specifically, however, LSTMs seem to perform higher than RNNs since LSTMs consistently gave the highest accuracy and sensitivity. A higher sensitivity means that there is a higher proportion of true positives to all positive cases. This is critical because it is important to minimize false negative occurrences because these are the most detrimental when considering the practical application of this. Additionally, having the highest accuracy is also important since the prediction model should give the most consistent predictions in relation with the actual occurrence of seizures.

The results show overall that the deep learning algorithms are successful in predicting seizures with a precision of 0.876 (RNN), accuracy of 0.804 (LSTM), specificity of 0.999 (CNN), and sensitivity of 0.642 (LSTM). RNN gave the highest precision of 0.876 for the 20 minutes interval in seizure prediction. LSTM outperformed RNN by resulting in a higher accuracy of 0.804 for 40 minutes interval. CNN resulted with the best specificity of +0.99 for all time interval models, with maximum value of 0.999 for 60 minutes interval. LR, and k-NN did not perform that well when it comes to predicting seizures in any of the performance metrics categories when compared with deep learning models. The results further highlight that as the prediction interval increases, the effectivity of this Prediction model decreases.

### 3.2 Channel Selection

10 channels were found to be most impactful in seizure forecasting out of the 23 original channels. To determine what channels independent component analysis had selected, each of the 16 channels were compared to each of the 10 ICA components using cosine similarity, which is a method of determining how similar two vectors are based on the cosine value of the angle formed between the two vectors. The channel that had the highest similarity score was subsequently determined to be the channel that ICA had selected. The 16 original channels were 'F7', 'T7', 'P7', 'F3', 'C3', 'P3', 'O1', 'F4', 'C4', 'P4', 'F8', 'T8', 'PO8', 'O2', 'FT9', and 'FT10', and the 10 channels selected by ICA were 'F7', 'T7', 'C3', 'P3', 'F4', 'C4', 'P4', 'F8', 'T8', and 'PO8'.

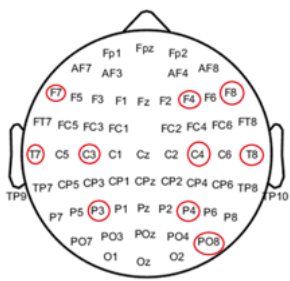

*Figure 15: Diagram showing the spatial locations of 10 channels selected by independent component analysis*

The channels seem to be scattered along the outer surface of the scalp, with none of the channels located in the center. One reason why these channels may have been selected is their ability to record activity from multiple brain regions. This is particularly useful since the seizure types present in the data are unspecified. It's possible that the seizure could be a generalized seizure, which occurs in all areas of the brain, and the placement of these channels would enable them to capture that activity. Additionally, the seizure could be a focal seizure, but since the channels are well-dispersed throughout the scalp except for the scalp's center, they would still be able to record activity in that specific region of the brain.

### 4. Conclusion and Future Research

In this study, a novel framework to predict seizures in Intractable Epileptogenic Brain Networks is proposed. The bandpass filtering and independent component analysis are proven to be effective in reducing the noise and separating out the artifacts from the dataset. In addition, ICA was able to successfully select 10 channels out of 23 which are the most influential in predicting seizures. The classification algorithms, specifically deep learning algorithms LSTM, RNN, CNN can find a relationship between the dataset's components and the electroencephalogram channels and perform better in measuring different performance metrics. Logistic Regression, and k-Nearest Neighbor did not perform well when it comes to predicting seizures in any of the performance metrics categories when compared with deep learning models. This research also shows that based on these models and their results, a maximum forecast time that can be achieved is one hour before the seizure occurrence, with the LSTM model giving the highest accuracy of approximately 80% at that time.

For future research, different training and testing datasets consisting of various age groups can be added to generalize the prediction model. A labeled dataset containing various types of seizures can be used to predict the effectiveness of this model. Different classification algorithms such as SVM, and transformers can be explored for different performance metrics in prediction. Prediction Models can be explored for longer time durations like 12 hours, 24 hours, 2 days, 7 days etc.

### 5. Acknowledgement

Special Thanks to Tanish Jain, Stanford University, Dr. Goldy Bansal, M.D, Overlake Hospital, WA, Michael Leung, Microsoft. Thank You to Boston Children Hospital, and MIT Research Lab for making the dataset available.